\documentclass{article}

\usepackage{arxiv}

\usepackage[utf8]{inputenc} 
\usepackage[T1]{fontenc}    
\usepackage{hyperref}       
\usepackage{url}            
\usepackage{booktabs}       
\usepackage{amsfonts}       
\usepackage{nicefrac}       
\usepackage{microtype}      
\usepackage{lipsum}
\usepackage{graphicx}
\usepackage{subcaption}
\usepackage{amsmath}
\usepackage{amsthm}

\title{Concept Drift Visualization of SVM with Shifting  Window}

\author{
  Honorius G\^ almeanu\\
  Department of Mathematics and Informatics \\ Transilvania University of Bra\c sov, Romania\\
  and\\
  FotoNation SRL - Tobii AB\\
  \texttt{galmeanu@unitbv.ro} \\
    \And
   R\u azvan Andonie\\
   Computer Science Department\\
   Central Washington University, Ellensburg, WA, USA\\
   and\\
   Transilvania University of Bra\c sov, Romania\\
   \texttt{razvan.andonie@cwu.edu} \\
}

\begin{document}

\maketitle

\begin{abstract}
In machine learning, concept drift is an evolution of information that invalidates the current data model. It happens when the statistical properties of the input data change over time in unforeseen ways. Concept drift detection is crucial when dealing with dynamically changing data. Its visualization can bring valuable insight into the data dynamics, especially for multidimensional data, and is related to visual knowledge discovery. We propose a novel visualization model based on parallel coordinates, denoted as parallel histograms through time. Our model represents histograms of feature distributions for successive time-shifted windows. The drift is shown as variations of these histograms, obtained by connecting the means of the distribution for successive time windows. We show how these diagrams can be used to explain the decision made by the machine learning model in choosing the drift point. By isolating the drift at the edges of successive time windows, there will be none (or reduced) drift within the adjacent windows. We illustrate this concept on both synthetic and real datasets. In our experiments, we use an incremental/decremental SVM with shifting window, introduced by us in previous work. With our proposed technique, in addition to detect the presence of concept drift, we can also depict it. This information can be further used to explain the change. mental results, opening the possibility for further investigations.
\end{abstract}


\section{Introduction}
The problem of underlying distribution change for the data a machine learning model is trained on is coined by the term {\it concept drift}. The real challenge is to recognize the difference between small data variations caused by noise versus data generated from non-stationary distributions \cite{IEEEexample:Gama2010}. To approach the problem of concept drift, several paradigms were proposed in the past years, including concept drift with shifting window. A comprehensive compilation can be found in \cite{IEEEexample:Lu2019}. 

Visualization of concept drift is an important problem from multiple perspectives: \emph{i)} in the early phase of dataset exploration it reveals the presence of concept drift; \emph{ii)} stationary character of the distribution can be assessed visually; \emph{iii)} concept drift can be "zoomed in", by examining a subset of the data; \emph{iv)} last but most important, it allows us to depict information that could be used to interpret or explain the decisions of the model, and this is related to Visual Knowledge Discovery (VKD) \cite{Kovalerchuk2024}. So far, visualization of concept drift was scarcely covered in literature \cite{IEEEexample:Pratt2003, IEEEexample:Few2009, IEEEexample:Webb2018, IEEEexample:Schneider2020, IEEEexample:Wang2020, IEEEexample:Hinder2023}. This gave us one motivation for the current work. We would like to explore the implications and applications of concept drift visualization in VKD.  

Concept drift can be detected by many statistical and machine learning tools, including Support Vector Machine (SVM). In previous work, we introduced an incremental-decremental SVM with adaptive window determined by statistical tests \cite{IEEEexample:Galmeanu2021}. We also used sample weighting within the shifting window in a follow-up of this work \cite{IEEEexample:Galmeanu2022}. The models were both targeting the concept drift detection problem. However, one aspect was not discussed: how can we explain the decisions made by the model, for instance when dropping samples from the training data. This aspect, related to explainable AI, was left unanswered and is the second motivation for our current research.

The closest result we could find to fit our needs is \cite{IEEEexample:Pratt2003}, which uses the visualization of parallel histograms. The time evolution of binned feature values are covered in \cite{IEEEexample:Schneider2020}. However, we could not find results regarding the evolution in time of feature histograms and their statistics. 

Our main contribution is the Parallel Histograms through Time (PHT) model. It combines visualizations of parallel feature histograms for consecutive time shifting windows - either disjoint or time-overlapped. We introduce the per-class visualization of these parallel histograms, depicting the class specific histograms for all visualized features, and tracing their associated mean as the window is shifting. We also propose a meta-algorithm to discover the points of drift using these visual representations.

The paper is structured as follows. The related work is presented in Section \ref{related}, with a focus on concept drift with SVM shifting windows and concept drift visualization models. Section \ref{method} introduces our PHT visualization model. Section \ref{experiments} presents several experiments performed on our visualization model using the incremental-decremental SVM with shifting window while proposing a meta-algorithm for visual discovery of concept drift. Section \ref{conclusion} includes final remarks  regarding the advantages of the proposed method.

\section{Related Work} \label{related}
To make the paper self-contained, we summarize in this section some results related to \emph{a)} concept drift detection using SVM  with shifting window,  and \emph{b)} concept drift visualization techniques.

\subsection{Concept drift with SVM shifting window} \label{background1}

A natural concept drift handling technique is based on instance selection: a window shifts over recently arrived instances and the learned concepts are used for prediction in the immediate future only \cite{IEEEexample:Tsymbal2004}. The forgetting mechanism can be extended by weighting the samples in a time window with fixed or variable size. During the last years, various concept drift models with shifting window were proposed (see:  \cite{IEEEexample:Iwashita2019} and \cite{IEEEexample:Lu2019}).


\begin{figure*}[!htb]
\centering
\begin{subfigure}{0.9\textwidth}
   \includegraphics[width=1\linewidth, height=0.2\textheight]{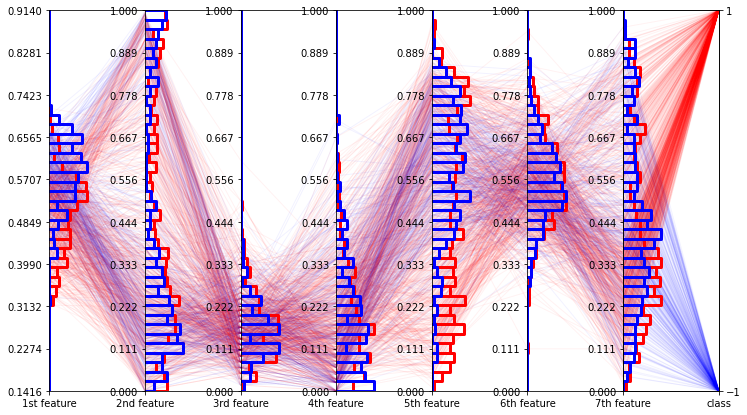}
   \caption{}
   \label{fig:1_1} 
\end{subfigure}
\vfill
\begin{subfigure}{0.9\textwidth}
   \includegraphics[width=1\linewidth, height=0.2\textheight]{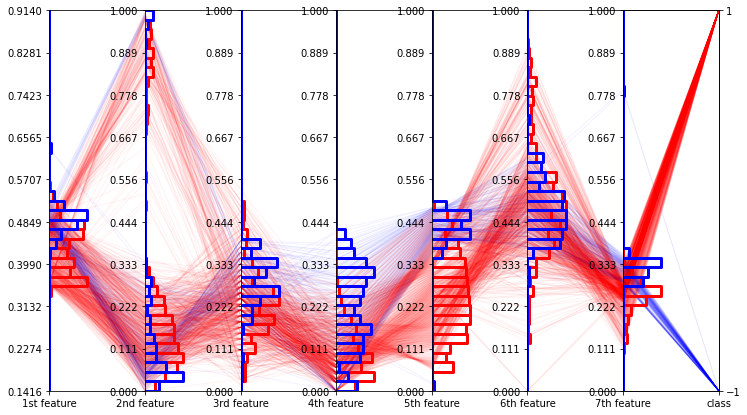}
   \caption{}
   \label{fig:1_2} 
\end{subfigure}
\caption{The seven most representative features of the CIRCLES dataset\cite{IEEEexample:Bifet2006} are represented. Parallel histograms consist of a parallel coordinate system where each vertical axis also contains a representation of the probability density (more often just a histogram). The height of each bin is relative to the height of the highest bin of that axis, and thus the histograms are not comparable among themselves. Plots in (a) and (b) represent the same dataset, but different time windows. The per-feature distributions are different in (a) vs. (b), revealing a concept drift. Each histogram uses 40 bins.}
\label{fig:1}
\end{figure*}

Concept drift detection with an incrementally trained SVM was discussed in \cite{IEEEexample:Syed1999}. The main idea was to discard, at each training step, all previous data except for the support vectors. An improved version was later proposed in \cite{IEEEexample:Ruping2001}. A SVM classification model beyond the learned label space was given in \cite{IEEEexample:ZareMoodi2016}. In this case, novel classes may emerge while processing the data stream - a challenging aspect of concept drift. More recently, \cite{IEEEexample:Jain2022} proposed the usage of shifting window combined with K-Means clustering to analyze drift, reducing the data size and upgrading the training dataset. The authors used a SVM to detect anomalies and statistical tests to initiate model retraining.

In order to deal with samples accumulation, various incremental strategies were proposed. In \cite{IEEEexample:Chitrakar2014}, an Improved Concentric Circles method was used. The points within the ring are kept for further training while the others are discarded. This specific ring is defined by two circles. The inner circle is determined by the class center and the nearest support vector, while the outer circle is tangent to the hyperplane. This blind heuristic could render inapplicable in certain cases; for example, in abrupt drift where classes change labels instantly, there is no justification for keeping any past samples. Another approach is  \cite{IEEEexample:Chen2019}, where an adaptive SVM speeds up the incremental updates using pre-calculated information and an incremental matrix update. 

\begin{figure*}[!htb]
\centering
\includegraphics[width=0.9\textwidth, height=0.25\textheight]{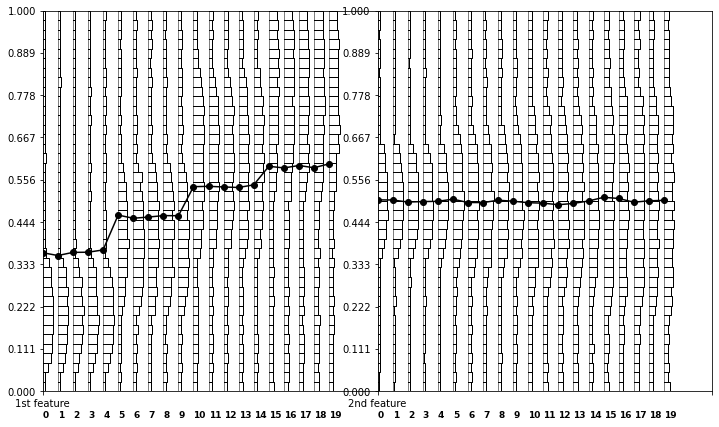}
\caption{PHT for a dataset with two features. Each feature is represented over a period of 20 disjoint time windows. For each time window, we represent its histogram and the associated mean. The segments that join these successive means indicate the presence of drift. We observe that the drift is more pronounced in the first feature.}
\label{fig:2}
\end{figure*}

\begin{figure*}[!htb]
\centering
\includegraphics[width=0.9\linewidth, height=0.25\textheight]{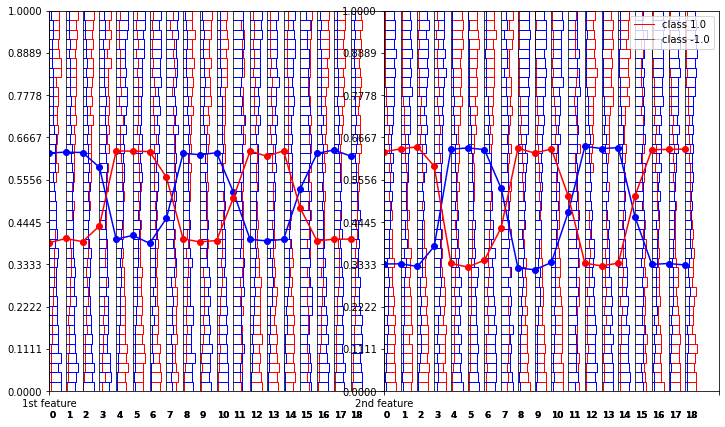}
\caption{Representation of the SINE1 dataset for 19 consecutive windows of 5,200 samples. The per class histograms and the means indicate the presence of sudden drift in both features.}
\label{fig:3}
\end{figure*}

Instance weighting and learning windows of variable length are used in \cite{IEEEexample:Klinkenberg2000, IEEEexample:Klinkenberg2004}. SVMs are used to recognize and handle the concept drift. At each training step, the model builds several SVM models using different window sizes, then selects the one minimizing the error estimate. 

In \cite{IEEEexample:Galmeanu2008, IEEEexample:Galmeanu2008b}, we extended the Cauwenberghs and Poggio's incremental/decremental SVM \cite{IEEEexample:Cauwenberghs2000}. Based on these results, we introduced an adaptive incremental/decremental SVM able to update the size of the shifting window leveraging the Hoeffding statistical test \cite{IEEEexample:Galmeanu2021}. We do not assume a fixed distribution; we use the statistical test to verify that the current sample belongs to the distribution described by the previously learned samples. If not, we start discarding the samples beginning with the least recent one, and recompute the distribution until the test becomes positive. As a generalization, in \cite{IEEEexample:Galmeanu2022} we proposed a weighted incremental/decremental SVM where the importance of each of the data sample in the shifting window can vary.

\subsection{Concept drift visualization} \label{background2}


Concept drift visualization refers to the ability of performing  descriptive analysis for data with time-changing statistical properties. Time-varying data like time-series does not necessarily imply concept drift. Therefore, concept drift visualization is more specific than visualization of time-varying data.

Some of the previous techniques used for concept drift visualization are general visualization tools like line charts, scatter plots, parallel coordinates, and heatmaps. For an overview of concept drift visualization techniques we refer to \cite{IEEEexample:Wang2020}.

One of the first successful works of this kind is presented in \cite{IEEEexample:Pratt2003}, and it coins the term of "brushed" parallel histograms. It begins by displaying a parallel coordinates graph, where the dimensions of the datasets are represented as parallel equally-spaced vertical axes. A sample represented in parallel coordinates system is a succession of line segments joining points on each of these parallel axes. A diagram of parallel histograms has histograms superimposed on each axis (Fig. \ref{fig:1}). The "brushing" implies that the user will select a range of values on an axis. The values included in the brushing are highlighted with color, and thus the user can view the inked values for all of the features on all axes simultaneously. This technique allows for the visualization of correlations between different dimensions but marginal feature distributions may be completely oblivious to class distribution changes. 

A more recent article \cite{IEEEexample:Webb2018} presents quantitative concept drift mapping techniques and their visualization. The authors start from the definition of concept as a probability distribution \cite{IEEEexample:Gama2014}, and then measure the total variation distance between two moments in time $t$ and $u$, as given by equation (\ref{eq:1}), where $Z$ represents a vector of random variables:

\begin{equation}\label{eq:1}
\sigma_{t,u}(Z) = \frac{1}{2}\sum\limits_{\overline{z} \in dom(Z)} |P_t(\overline{z}) - P_u(\overline{z})| 
\end{equation}

The probability distribution of the two time intervals is estimated using maximum likelihood. The variation in time of the covariate drift and of the conditional marginal covariate drift are investigated. Heatmaps showing the  distribution of drifts between pairs of dataset attributes are also visualized. An assumption is that the distribution within each of the compared time intervals does not change, which is not always the case. 

ConceptExplorer \cite{IEEEexample:Wang2020} is a visual detection scheme for discovering concept drift among multiple sourced time series based on prediction models. The concept drift is detected by observing the performance drop of a trained classifier. The distribution of correct predictions in a shifting window is assimilated to a binomial distribution. Drift is detected by observing the classifier's error rate to be in the upper tail of the 99 percentile. Interesting time segments are selected and shown in a timeline navigator view. From there, the user observes the performance drop in the accuracy fluctuation chart. Parameters of the model are selected and their PCA representation is projected on a 2D plane. This allows for visualization of the similarity described by the predictor models in the presence of data drift. Eventually, the concept explanation view shows a correlation matrix between pairs of selected attributes. The attributes are expanded into multiple intervals. This allows multiple rows (columns) for that attribute. For each intersection square, the difference and the sum between the number of records with positive labels and those with negative labels is computed, and their ratio is shown as a heatmap. Observing that the correlation matrix is symmetric, the authors represent the two symmetric off-diagonal cells using two different time intervals. It is expected that the concept drift is to be observed as a difference of color between the two cells corresponding to attributes' correlations in two different time intervals. However, the localisation is poor, as this depends on the granularity of the defined time intervals.

DataShiftExplorer \cite{IEEEexample:Schneider2020} is an interactive visual technique for identification and analysis of the change in multidimensional data distributions. It starts from showing superimposed per-feature density plots that compare training data distributions with unseen data distributions. The core feature is the data-shift pattern diagram. The normalized difference of counts between data bins per feature is plotted as dots of variable size and color intensity. The binned features are represented through time in a 2D diagram. The binned feature values for every multi-dimensional data record are joined, producing a parallel coordinates plot.

Another way of visualizing concept drift is to train specific prediction models. These allow for the extraction and visualisation of the features mostly associated with the drift, based on feature importance \cite{IEEEexample:Hinder2023}. Once the features are isolated, heat maps are produced, visualizing the feature distribution over time or depicting pairwise correlation of features at different time moments. 

The techniques used so far in the literature and examined above are not sufficient, since they do not describe the time dynamics of the features. We found the literature two useful representations: either \emph{i)} individual features' histograms represented for a specific time window, or \emph{ii)} the variation in time of a single feature characteristic (e.g. the mean). We would like to visualize, in a single descriptive diagram, both changes of histograms vs. time and also the variation of features' associated characteristics.

\section{Parallel Histograms through Time} \label{method}
This section introduces our PHT visualization framework. 
Almost all of the previous works make use of some sort of feature's probability density in a time-unfolding diagram. Inspired by \cite{IEEEexample:Pratt2003} and \cite{IEEEexample:Schneider2020}, we chose to enhance the parallel histograms system. 

The parallel coordinates system \cite{IEEEexample:Pratt2003}
consist of a set of vertical parallel lines, each one corresponding to a distinct feature of the dataset being analyzed. A vertical line represents the entire domain of a feature. A sample from the dataset 
is obtained by joining points on these axes associated with specific features' values. This is close to a time-series representation, with the  exception that time has no representation here: the vertical axes are not ordered in time. On this coordinate system, one can superimpose, for each vertical axis, the histogram associated with the probability distribution of that dimension \cite{IEEEexample:Pratt2003}. Fig. \ref{fig:1_1} depicts the parallel histograms for a certain time window of the Forest Covertype dataset \cite{IEEEexample:Blackard2000}. For an ulterior time window of the same size, the distributions of all of the ten features displayed change (Fig. \ref{fig:1_2}). To visualize many more moments in time, this representation turns out to be inappropriate, and we have to think of something new.

\begin{figure*}[bht]
\centering
\begin{subfigure}{0.9\textwidth}
   \includegraphics[width=1\linewidth, height=0.25\textheight]{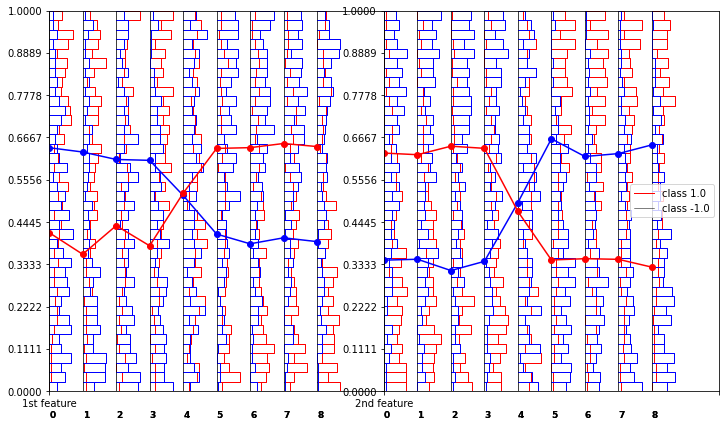}
   \caption{}
   \label{fig:4_1} 
\end{subfigure}
\vfill
\begin{subfigure}{0.9\textwidth}
   \includegraphics[width=1\linewidth, height=0.25\textheight]{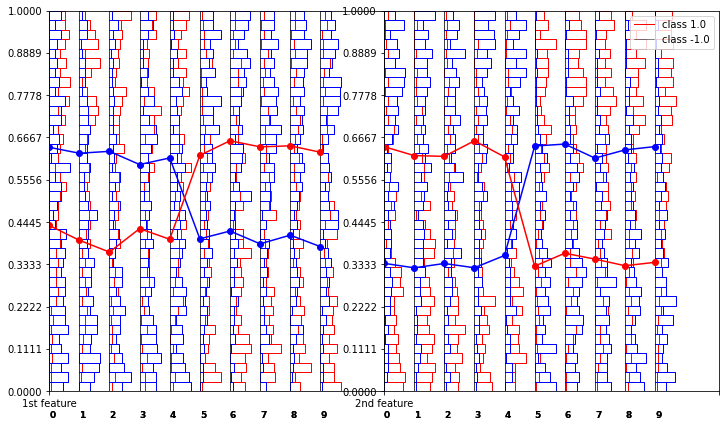}
   \caption{}
   \label{fig:4_2} 
\end{subfigure}
\caption{PHT representation for consecutive windows for the SINE1 dataset. The window size in (a) was chosen such that the concept drift changes are present inside the middle of a window and mimics the real scenario where drift position is not known apriori. In (b), the drift was identified by the SVM and we were able to isolate it in-between windows.}
\label{fig:4}
\end{figure*}

For this reason, we propose the PHT representation, that combines the parallel histograms with a time-series representation (Fig. \ref{fig:2}). We retain the parallel histograms, but we skip plotting the dataset samples, as this will clutter the plot. As we analyze the data stream, we consider consecutive disjoint time windows of constant size. For each of the windows we compute the histograms associated with each feature. Then, the histograms are depicted in the same subplot of their associated feature. 

In Fig. \ref{fig:2}, 20 separate time windows and only two features are considered. For each histogram we also mark the mean value of the feature. Joining the means together leads to a curve. For the ideal case when we have no drift, the curve becomes a horizontal line. Anything different than a horizontal line clearly indicates a concept drift. We can observe in Fig. \ref{fig:2} that the drift is more pronounced for the first feature. The second feature seems to follow the same distribution; however, the histogram becomes wider, indicating a larger standard deviation and thus a distribution change. In certain cases, the method fails to show a drift for a specific feature, whereas the drift is present if classes are investigated individually. From this perspective, the method is similar to a naive Bayes classifier - it considers only the independent and marginal probability distributions, assuming that the the features are statistically independent. Since it only illustrates the first order statistics of the features, our model does not generate false positives, but is rather prone to exhibit false negatives.

Another limitation comes from the number of represented features. Usually the visualization becomes cluttered if one represents more than a dozen features. Since the horizontal axis is used to represent both features and for each feature a number of shifting windows is shown, a balance between the number of features and the number of consecutive shifting windows should be maintained. One may represent only the most important features, considering a custom ranking. 

\section{Experiments and drift identification} \label{experiments}
We introduce the following meta-algorithm to locate the points of concept drift:

\begin{figure*}[!htb]
\centering
\includegraphics[width=1\linewidth, height=0.3\textheight]{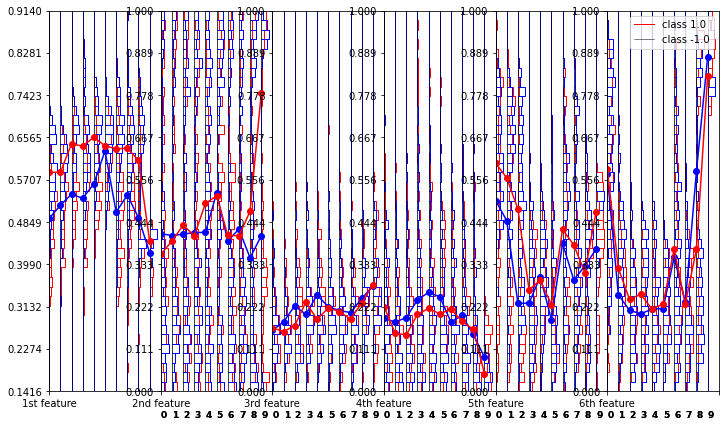}
\caption{PHT for the Forest Covertype dataset. Six of the ten continuous features were used. We show here the first 10 consecutive windows of 500 samples each. As the diagram shows, steady drift is present in the features.}
\label{fig:5}
\end{figure*}

\begin{enumerate}
    \item Represent the dataset using the parallel histograms:
    \begin{enumerate}
    \item Remove the categorical features with no variability (for example, features maintaining the same value through all the dataset - they are irrelevant pertaining the concept drift).
    \item Remove the continuous features with zero or small variability (i.e., constant features or very narrow interval features).
    \end{enumerate}

    \item Represent the so-far filtered dataset using parallel histograms through time (PHT):
    \begin{enumerate}
    \item If there are too many features to be represented, do partial representations with at most a dozen features and filter out the features that present small or no concept drift - their means line is almost horizontal, with slight or no variability; collect only the relevant features that manifest significant concept drift.
    \item From the remaining features, if they are still too many, select the ones that present the most variability.
    \end{enumerate}

    \item On the PHT diagram with the features filtered out through the previous steps:
    \begin{enumerate}
    \item Choose a sufficiently small window size, such that the means line looks horizontal on segments. This indicates that there is small or no concept drift on these segments. One would observe something similar with the situation in Fig. \ref{fig:4_1}, characterized by abrupt concept drift at certain windows. Relative stability is observed during windows 0 - 3 and 5 - 8. It could be that no matter how small the window size, the drift is still observable (e.g., Fig. \ref{fig:5}, characterized by continuous drift).
    \item To localize the abrupt drift, work on the window size where the drift appears - in our example, windows 3 - 5 in Fig. \ref{fig:4_1}. Change the window size until clear window cutting points appear, with no transition means, like the one in Fig. \ref{fig:4_2} - windows 4 and 5, although adjacent, shows an abrupt change of means, but with no intermediate point like window 4 in Fig. \ref{fig:4_1} above. Usually this is observed simultaneously in many features.
    \end{enumerate}
\end{enumerate}

We train our incremental/decremental SVM with adaptive shifting window, introduced in  \cite{IEEEexample:Galmeanu2021}, on several datasets. This SVM  has the property that, once the concept drift is detected, the current shifting window is adjusted by omitting the earliest samples, until no drift is detected on the remaining ones. Our implementation can be accessed and replicated through the adaptive incremental-decremental SVM code provided on GitHub\footnote{\url{https://github.com/hash2100/aidsvm}}. The code for the diagrams is also made available\footnote{\url{https://github.com/hash2100/iv-concept-drift}}.

The CIRCLES dataset \cite{IEEEexample:Bifet2006} is a set with gradual drift: there are two attributes $x$ and $y$ uniformly distributed in the interval $[0, 1]$. There is a circle defined, of given center and radius. The positive class samples are inside the circle, whereas the ones on the outside are labeled as negative. Concept drift occurs when the classification function, that is the circle parameters, change.
This happens every 25,000 samples. We illustrate the concept drift in Fig. \ref{fig:1}. 

In Fig. \ref{fig:2}, each represented histogram corresponds to a window of 5,000 samples. The windows do not overlap - they are consecutive. The representation shows how the sample mean abruptly changes after five consecutive windows.

Another frequently used dataset in the literature is the SINE1 synthetic set \cite{IEEEexample:Pesaranghader2018}. It has two classes, 100,000 samples and abrupt concept drifts. Additionally, 10\% white noise was added to the data. The dataset has only two attributes, $(x_a, x_b)$ uniformly distributed in $[0, 1]$. A point $x_b < sin(x_a)$ is classified as belonging to one class, with the rest belonging to the other class. At every 20,000 instances an abrupt drift occurs and the classification is reversed. We visualize the dataset using PHT in Fig. \ref{fig:3}. We represent consecutive windows of 5,200 samples each. The window size corresponds to the situation where we do not know a priori the position of the drift. The two classes are represented with different colors. We can observe how the drift occurs:  the classes switch their labels. Our method can be used to identify the drift point. For example, in Fig. \ref{fig:4_1} we represent nine consecutive windows of 500 samples, starting from sample 17,800. The drift occurs right at the middle of the fifth window. Using the SVM classifier, the drift is detected to occur at sample 20,050. We can correct the representation by depicting ten consecutive windows of 500 samples each, starting from position 17,550. This way, the abrupt drift occurs right after the fifth window, and it can be seen in Fig. \ref{fig:4_2}.

By following the steps of the meta-algorithm above we located and revealed the points of concept drift in SINE1 dataset. These were confirmed by the window size profiles created by the incremental-decremental SVM classifier with adaptive window - they are the points where the window size suddenly decreases. 

The Forest Covertype dataset \cite{IEEEexample:Blackard2000} describes evolution of forest coverage in 30 $\times$ 30 meter cells, provided by the US Forest Service (USFS). The dataset has 54 attributes and 7 classes. We only use the two most represented classes and six least trivial features of the 10 continuous features, in total 495,141 samples. We show only the first 5,000 samples in consecutive non-overlapping windows of 500 samples in Fig. \ref{fig:5}. A steady concept drift is visible. This is confirmed by the SVM classifier, as the size of its shifting window keeps fluctuating \cite{IEEEexample:Galmeanu2021}.

\section{Conclusion} \label{conclusion}
We introduced the PHT model, a visual tool meant to reveal the occurrence of concept drift within datasets. This is depicted as changes within the histogram and the associated sample mean for consecutive time windows. The tool helps in visualizing and also locating the points of sudden change, allowing for simultaneously visualization of the evolution for about a dozen features. Used in conjunction with an online classifier like the incremental/decremental SVM, it leverages the classifier with a visual explanation of detected concept drift in streaming data. Without this visualization, the classifier can detect but not show the change. This visualization may be used to further justify or explain the change.


\end{document}